\documentclass[10pt,twocolumn,letterpaper]{article}

\usepackage{iccv}              
\usepackage{multirow}

\definecolor{iccvblue}{rgb}{0.21,0.49,0.74}
\usepackage[pagebackref,breaklinks,colorlinks,allcolors=iccvblue]{hyperref}

\def\paperID{14220} 
\def\confName{ICCV}
\def\confYear{2025}

\title{GenM\(^3\): Generative Pretrained Multi-path Motion Model \\for Text Conditional Human Motion Generation}

\author{First Author\\
Institution1\\
Institution1 address\\
{\tt\small firstauthor@i1.org}
\and
Second Author\\
Institution2\\
First line of institution2 address\\
{\tt\small secondauthor@i2.org}
}

\author{%
  Junyu Shi\textsuperscript{1}, 
  Lijiang Liu\textsuperscript{1}, 
  Yong Sun\textsuperscript{1}, 
  Zhiyuan Zhang\textsuperscript{1}, 
  Jinni Zhou\textsuperscript{1}, 
  Qiang Nie\textsuperscript{1}\thanks{Corresponding author}\\
  \textsuperscript{1}The Hong Kong University of Science and Technology (Guangzhou)\\
  {\footnotesize \texttt{\{jshi890, lliu135, ysun691, zzhang400\}@connect.hkust-gz.edu.cn}\quad
  \texttt{\{eejinni, qiangnie\}@hkust-gz.edu.cn}}
}

\begin{document}
\maketitle
\begin{abstract}
Scaling up motion datasets is crucial to enhance motion generation capabilities. However, training on large-scale multi-source datasets introduces data heterogeneity challenges due to variations in motion content. To address this, we propose Generative Pretrained Multi-path Motion Model (GenM\(^3\)), a comprehensive framework designed to learn unified motion representations. GenM\(^3\) comprises two components: 1) a Multi-Expert VQ-VAE (MEVQ-VAE) that adapts to different dataset distributions to learn a unified discrete motion representation, and 2) a Multi-path Motion Transformer (MMT) that improves intra-modal representations by using separate modality-specific pathways, each with densely activated experts to accommodate variations within that modality, and improves inter-modal alignment by the text-motion shared pathway. To enable large-scale training, we integrate and unify 11 high-quality motion datasets (approximately 220 hours of motion data) and augment it with textual annotations (nearly 10,000 motion sequences labeled by a large language model and 300+ by human experts). After training on our integrated dataset, GenM\(^3\) achieves a state-of-the-art FID of 0.035 on the HumanML3D benchmark, surpassing state-of-the-art methods by a large margin. It also demonstrates strong zero-shot generalization on IDEA400 dataset, highlighting its effectiveness and adaptability across diverse motion scenarios.
\vspace{-1cm}
\end{abstract}    
\section{Introduction}
\label{sec:intro}

\begin{figure}[t!]
    \centering
    \includegraphics[width=0.45\textwidth]{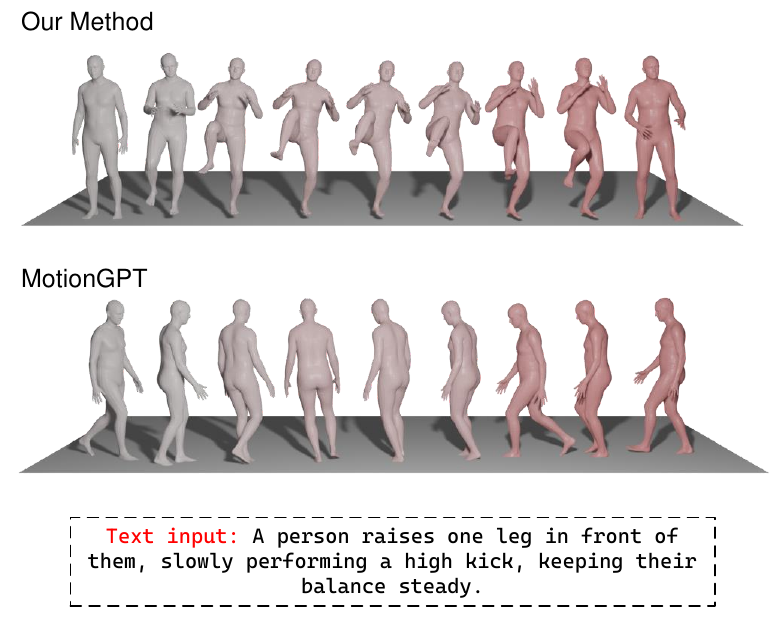}
    \vspace{-0.25cm}
    \caption{Due to the limited motion and text in the dataset, MotionGPT~\cite{jiang2023motiongpt} is unable to respond to this relatively simple text input. We expanded the dataset in both motion and text modalities and pre-trained our model on this large-scale dataset. This enabled our model, GenM\(^3\), to generate high-precision human motions based on diversity descriptions.} 
    \label{fig:comparison}
    \vspace{-0.25cm}
\end{figure}

Generating diverse and accurate human motion from textual descriptions has emerged as a vital area in generative computer vision~\cite{guo2022humanml3d, petrovich2022temos, tevet2022motionclip, chen2023executing, yuan2023physdiff, tevet2023human, harvey2020robust}, which has broad applications across animation, virtual reality, and robotics. 

Inspired by breakthroughs of pre-trained Large Language Models (LLMs) on large-scale datasets, recent research in human motion generation leverages larger models~\cite{liang2024omg} and expanded datasets~\cite{lin2023motionx, guo2022humanml3d} to produce more realistic and nuanced motion sequences. On the data side, datasets like HumanML3D~\cite{guo2022humanml3d} and Motion-X~\cite{lin2023motionx} have broadened available training samples. In terms of modeling, approaches like MotionGPT~\cite{jiang2023motiongpt} fine-tunes pre-trained LLMs on the HumanML3D dataset to jointly model motion tokens and text tokens. Building upon this, MotionChain~\cite{Jiang2025motionchain} extends the use of pretrained LLMs for multi-turn motion dialogue, enabling dynamic interactions over sequential motions. Another pioneering model, OMG~\cite{liang2024omg}, incorporates multiple motion datasets and utilizes a large-scale diffusion model to achieve strong zero-shot performance. These approaches underscore the promising role of pretrained LLMs and large-scale data integration in enhancing the realism of human motion generation.

Despite above advancements, several challenges persist in motion generation: 1) Heterogeneity in motion data distribution. The heterogeneity arising from differences in motion types, recording devices, and similar factors poses a challenge for joint training, as optimizing performance on one dataset may adversely affect the results on another dataset. 2) Lack of a dedicated pretrained motion backbone for comprehensive motion representations. While vision/language pretrained backbones have demonstrated robustness in image and text processing, efforts to adapt large language models to motion tasks~\cite{jiang2023motiongpt, Jiang2025motionchain, wu2024motionagent, wang2024motiongpt} are hindered by inherent structural and contextual differences between motion and text. Thus, a dedicated motion model pretrained on large-scale motion data are crucial for capturing motion data's unique representations. 3) Lack of high-quality, large-scale unified motion datasets. The high cost of capturing high-fidelity human motion results in smaller datasets compared to natural images. Moreover, Existing datasets are often limited to specific motion types (e.g., HumanML3D emphasizes self-motion, while HIMO focuses on human–object interactions), restricting generalizability across varied human activities. Figure \ref{fig:comparison} illustrates an example of motion generation. Despite the simple description, MotionGPT~\cite{jiang2023motiongpt} failed to grasp the textual meaning and produced almost irrelevant motion.

\begin{figure*}[t!]
\vspace{-0.5cm}
    \centering
    \includegraphics[width=0.97\textwidth]{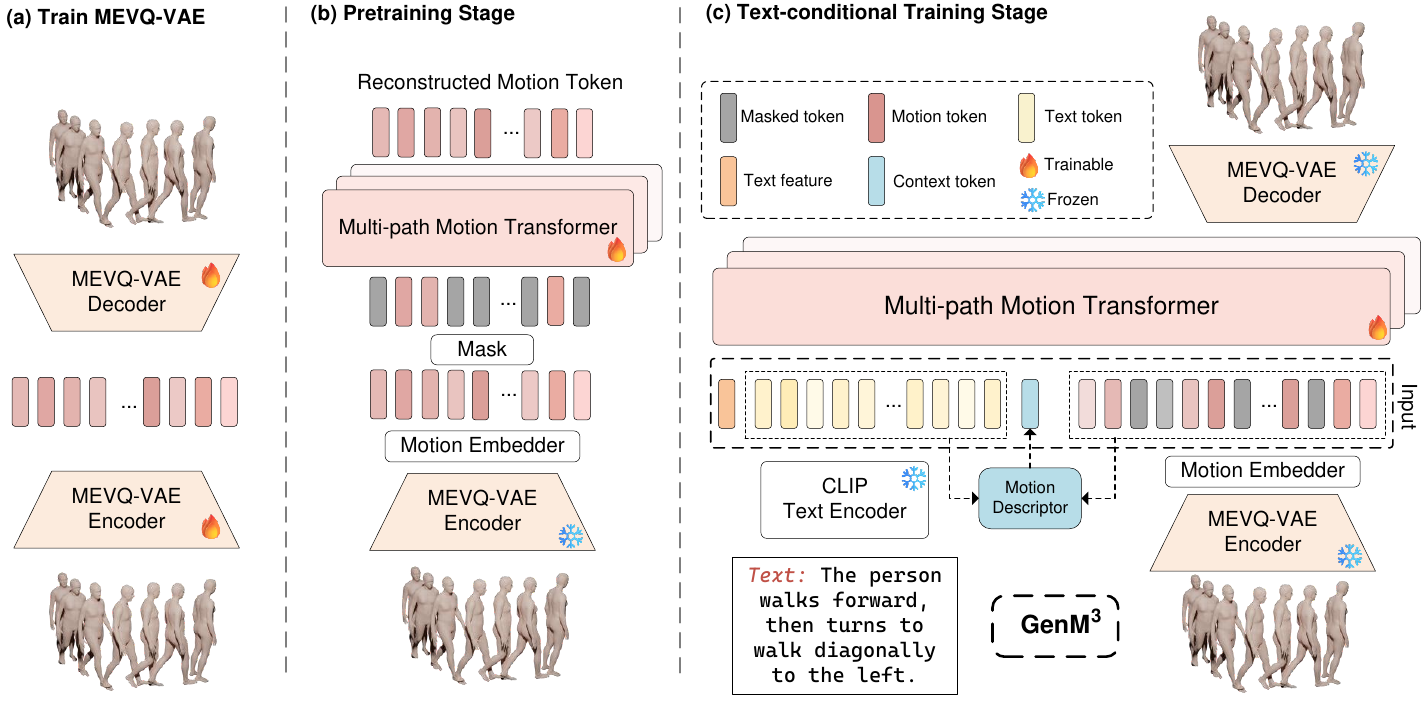}
    \caption{Overview of the GenM\(^3\) framework. The training of GenM\(^3\) consists of three steps: 1) Train MEVQ-VAE to extract discretized motion representations. MEVQ-VAE is frozen in the later steps. 2) Pre-train GenM\(^3\) on a large amount of motion modality data using self-reconstruction with masked tokens. 3) Perform text-conditional training on text-motion pairs. The processed text tokens are concatenated with the embedded discretized action tokens as input to the Multi-path Motion Transformer (MMT). The output of MMT is then fed into the Decoder of MEVQ-VAE, resulting in the generated motion sequence. Note that the computations inside GenM\(^3\) are length-independent. } 
    \label{fig:pipeline}
    \vspace{-0.5cm}
\end{figure*}

To overcome these challenges, in this paper, we propose Generative Pretrained Multi-path Motion Model (GenM\(^3\)), a framework designed to learn unified motion representations to address the challenge of data heterogeneity in large-scale, multi-source motion datasets. GenM\(^3\) comprises two novel designs: Multi-Expert VQ-VAE (MEVQ-VAE) and Multi-path Motion Transformer (MMT). As shown in Fig. \ref{fig:pipeline}, MEVQ-VAE discretizes continuous motion sequences through a multi-expert architecture that simultaneously activates all experts and adaptively adjusts each expert's contribution based on learned weights, capturing shared characteristics across datasets while preserving unique data attributes. Drawing inspiration from multimodal models such as VLMO~\cite{bao2022vlmo}, MMT utilizes separate modality-specific pathways, each with densely activated experts to adapt variations of data distribution within that modality. To facilitate cross-modal alignment, we also integrate a shared text-motion pathway that learns joint representations, bridging the gap between textual and motion semantics. That is, MMT employs three distinct pathways: one for the motion mode, one for the text mode, and a cross-modal pathway that handles both motion and text data. In addition, GenM\(^3\) integrates text tokens with motion tokens through a Motion Descriptor, producing context tokens representing overarching motion summaries, providing text-guided motion insights that enhance cross-modal alignment. The Multi-path Motion Transformer concurrently processes text tokens (including context tokens) and motion tokens, learning alignments between modalities. 

We integrate and unify 11 high-quality motion datasets to support GenM\(^3\)'s training and provide a large-scale motion dataset across multiple scenarios. We preprocess and align these datasets in terms of data structure and pose representation to match the benchmark dataset, HumanML3D~\cite{guo2022humanml3d}. After processing, the effective motion duration is approximately 220 hours, covering scenarios including single-person motion, two-person interactions, and human-object interactions. Notably, the HIMO~\cite{lv2025himo} dataset includes textual descriptions of interactions between upper limbs and objects, while the HuMMan~\cite{cai2022humman} dataset provides detailed textual annotations of joint changes during movement. Additionally, we enriched the IMHD~\cite{zhao2024imhoi} dataset by manually annotating full-body interactions with objects and generated segment-level descriptions for the BABEL~\cite{punnakkal2021babel} dataset using ChatGLM~\cite{glm2024chatglm}. These diverse textual descriptions extend HumanML3D's representation of human motion, enabling the model to respond effectively to a broader array of motion descriptions.

We pretrain GenM\(^3\) on all motion datasets using a masked completion strategy, which harnesses the model’s potential as a powerful motion representation tool by exposing it to diverse motion patterns. Subsequently, we conduct text-conditional generative training on datasets containing motion-text pairs, adapting GenM\(^3\)'s motion representation capability to text-driven generation. In summary, our contributions include the following:

\begin{itemize}
    \item We propose GenM\(^3\) to address data heterogeneity and facilitate cross-modal alignment during training with large-scale datasets. To this end, we integrate and unify 11 high-quality motion datasets (nearly 220 hours) across diverse scenarios, and enriching them with textual labels.
    \item In GenM\(^3\), we propose Multi-Expert VQ-VAE to learn a unified discrete motion representation, and a Multi-path Motion Transformer to adapt to text and motion variations and enhance text-motion alignment.
    \item GenM\(^3\) achieves 0.035 FID on HumanML3D dataset, significantly surpassing SoTA mathods, and shows superior zero-shot performance on IDEA400 dataset.
\end{itemize}

\section{Related Work}

\subsection{Text-Conditional Motion Generation}
Motion generation is a fundamental research area in computer vision and graphics, aiming to produce realistic and contextually appropriate human motions. Text-conditional motion synthesis, in particular, focuses on generating motion sequences based on textual descriptions. Early methods employ deterministic models, which could result in averaged or less dynamic motions due to the complex and stochastic nature of human movement. To capture a broader range of motion variations, researchers have turned to probabilistic models like Generative Adversarial Networks (GANs)~\cite{harvey2020robust, ghosh2021synthesis} and Variational Autoencoders (VAEs)~\cite{petrovich2022temos}, which help in modeling the inherent uncertainty and diversity in motion data. Diffusion models~\cite{du2023avatars, chen2023executing, yuan2023physdiff, tevet2023human, zhang2023finemogen, zhang2023remodiffuse, zhang2024motiondiffuse, xie2024omnicontrol, zhou2025emdm, wang2023fg} have also gained traction for their ability to generate high-quality motions by iteratively refining random noise into coherent motion sequences. More recently, some approaches~\cite{guo2022tm2t, du2023avatars, zhong2023attt2m, zhang2023generating, pinyoanuntapong2024bamm, guo2024momask, pinyoanuntapong2024mmm, gong2023tm2d} employ VQ-VAE to discretize motion sequences, followed by using a generative Transformer to produce motion tokens. Specifically, T2M-GPT~\cite{zhang2023generating} generate motion tokens in a next-token prediction paradigm, while MMM~\cite{pinyoanuntapong2024mmm} and MoMask~\cite{guo2024momask} utilize a masked completion approach: during training, they reconstruct randomly masked motion tokens, and in inference, they iteratively generate motions starting from a fully masked sequence.

\subsection{Large Model for Motion Generation}
The use of Large Model in motion generation has gained increasing attention due to their success in capturing complex, contextually rich dependencies within motion data. MotionGPT~\cite{jiang2023motiongpt} applied pretrained LLMs~\cite{raffel2020exploring} as the backbone for motion modeling. By discretizing motion sequences, motion tokens are treated as a new language, allowing joint modeling with text. Building on MotionGPT, MotionChain~\cite{Jiang2025motionchain} introduced a multi-turn motion generation mechanism that enables more natural interactions. Although these approaches leverage the powerful representational capabilities of pretrained LLMs, there exists a significant domain gap between motion and text data, which can impact the reliability of the generated motions.
On the other hand, the scale of the dataset is a critical factor for successfully training large motion models. HumanML3D~\cite{guo2022humanml3d}, which labels textual annotations to AMASS~\cite{mahmood2019amass} motion data, provides a benchmark dataset for motion generation. Motion-X~\cite{lin2023motionx} expands on this by using an automated labeling pipeline to extract a large number of motion sequences from video and generate corresponding text descriptions. However, the motion data in Motion-X contains significant noise, which limits its utility for high-precision motion generation. The OMG~\cite{liang2024omg} model integrates a wide range of motion datasets and trains various sizes of diffusion-based models for motion generation, pushing forward the capacity of large-scale motion modeling. Despite the success of these methods, substantial room remains for improvement in multi-dataset joint training and large-scale motion representation models within the field of motion generation. In this paper, we aim to center the analysis on motion data, transferring the scale law of datasets to the action domain, in order to mitigate the impact of the domain gap of large language models (LLMs) on motion data, and provide community with a comprehensive motion backbone.
\section{Method}

\begin{figure}[t!]
\vspace{-0.5cm}
    \centering
    \includegraphics[width=0.47\textwidth]{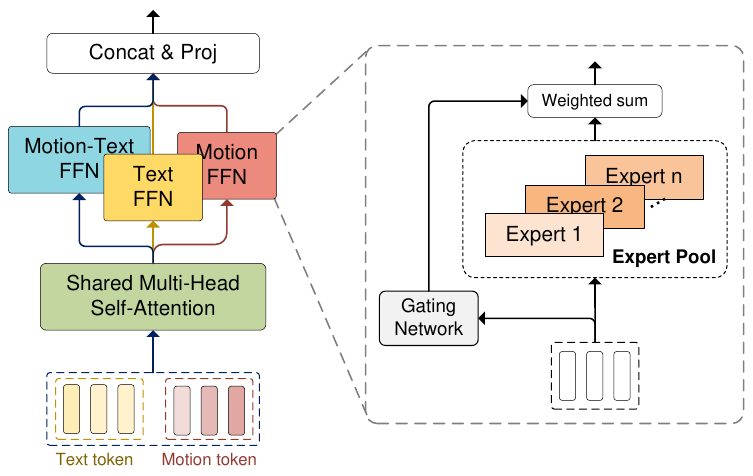}
    \caption{Architecture of multiway transformers in GenM\(^3\). The text, motion, and motion-text (cross-modal) branches handle their respective tokens, and then their outputs are aggregated at the output end. Each branch is equipped with an expert pool.} 
    \label{fig:mmtrans}
    \vspace{-0.5cm}
\end{figure}

\subsection{Model Architecture}
Our proposed GenM\(^3\) framework comprises two main components: the Multi-Expert VQ-VAE (MEVQ-VAE) and the Multi-path Motion Transformer (MMT). The MEVQ-VAE is designed to discretize continuous motion data while adapting to distributional differences across datasets. Building on these discrete tokens, the MMT models both motion and textual tokens in a unified manner, leveraging a multi-branch strategy to enhance intra-modal and inter-modal representational capacity. 

\subsubsection{Multi-Expert VQ-VAE}
The Multi-Expert VQ-VAE (MEVQ-VAE) is an extension of the original VQ-VAE framework, designed to discretize continuous motion data while handling variations across multiple datasets. As shown in Fig. \ref{fig:pipeline} (a), given a motion sequence \( \mathbf{X} = \left[ x_1, x_2, ..., x_T \right] \) with \( x_t \in \mathbb{R}^d \), where \( T \) is the number of frames and \( d \) is the dimensionality of each frame, the MEVQ-VAE encodes the sequence through the encoder \( \mathcal{E} \) and recover the complete motion sequence by the decoder \( \mathcal{D} \). A shared learnable codebook is employed, which contains \( K \) discrete codes \( \mathbf{C} = {c_k}_{k=1}^K \) with \( c_k \in \mathbb{R}^{d_c} \), where \( d_c \) is the dimensionality of each code. Each latent feature \( \mathbf{Z} = \mathcal{E}\left( \mathbf{X} \right) = \left[ z_1, z_2, ..., z_{T/l} \right] \), with \( z_i \in \mathbb{R}^{d_c} \) and \( l \) being the downsampling rate, is quantized by selecting the closest code from \( \mathbf{C} \).

The encoder \( \mathcal{E} \) and decoder \( \mathcal{D} \) is composed of multiple blocks, each containing three standard 1D convolutional layers followed by a specialized 1D convolutional layer with \( e_{q}\) experts. In this expert layer, all experts are activated simultaneously to capture diverse motion features, which can be represented as a weighted combination of each expert's output, formulated as:
\begin{align}
    y=\sum_{i=1}^{e_{q}} w_{i} \cdot \operatorname{Conv}_{i}(x)
\end{align}
where \( \operatorname{Conv}_{i}(x) \) represents the output from the \( i \)-th expert's convolutional filter, and \( w_i \) is the learnable weight associated with each expert.

\subsubsection{Multi-path Motion Transformer}
The Multi-path Motion Transformer (MMT) framework consists of two modules: the Motion Descriptor and a Multi-path Transformer network. The Motion Descriptor is responsible for summarizing motion patterns, converting them into context tokens that provide high-level representations of the motion sequence. The Transformer network then processes three types of tokens: text tokens, context tokens (considered as a type of text token), and motion tokens.

\paragraph{Motion Descriptor}
Inspired by LLaMA-VID's approach~\cite{li2024llama} to text information compression, we design the Motion Descriptor to enhance motion representation. After processing the motion sequence \( \mathbf{X} \) through MEVQ-VAE, it is mapped to a sequence of indices \( \mathbf{S} = \left[ s_1, s_2, ..., s_{T/l}, s_{End} \right] \) within a learned codebook. Passing \( \mathbf{S} \) through a motion embedder \( \operatorname{Emb}_{m} \), we obtain the motion embedding \( \mathbf{E}_m \). Simultaneously, the input text is encoded by a CLIP encoder to generate both a text embedding \( \mathbf{E}_t \) and a global text feature \( \mathbf{e}_t \).

To construct a contextual summary embedding \( \mathbf{E}_{ctx} \) for motion, we aggregate motion features by scoring their response to the text query:
\begin{align}
    \mathbf{E}_{ctx} = \operatorname{mean}\left( \operatorname{softmax}\left( \mathbf{E}_m \mathbf{E}_t \right) \mathbf{E}_t \right)
\end{align}
This selective aggregation captures a high-level motion summary within \( \mathbf{E}_{ctx} \), which integrates both motion and textual perspectives and serves as an enriched contextual input within the textual branch of MMT.

\paragraph{Multi-path Transformer}
We concatenate text features \( \mathbf{e}_t \), text embeddings \( \mathbf{E}_t \), context embeddings \( \mathbf{E}_{ctx} \), and motion embeddings \( \mathbf{E}_m \), adding positional encoding to form the input for our multi-path Transformer. The first half of the Transformer follows the standard Transformer architecture, utilizing self-attention and FeedForward layers to compute cross-modal attention:
\begin{align}
    \mathbf{E}_u = \operatorname{concat}\left( \mathbf{e}_t, \mathbf{E}_t, \mathbf{E}_{ctx}, \mathbf{E}_m \right) + \mathbf{P} \\
    \mathbf{Y} = \operatorname{FFN}\left( \operatorname{Attention} \left( \mathbf{E}_u \right) \right)
\end{align}
where \( \mathbf{P} \) and \( \operatorname{FFN} \) represent the Position Encoding and FeedForward Network, respectively.

As shown in Fig. \ref{fig:mmtrans}, In the whole second half of the model, we introduce specialized pathways within the FeedForward layer. These pathways include a motion pathway, a text pathway, and a cross-modal shared pathway, which are parallel. To further adapt to varying data distributions and diverse textual descriptions across datasets, we incorporate a multi-expert strategy within each pathway. Mathematically, each pathway \( p \) consists of multiple experts, and for each expert \( i \), we compute:
\begin{align}
    \mathbb{E}_{p,i}\left( x \right) = \mathbf{W}_{p,i}^{2} \sigma \left( \mathbf{W}_{p,i}^1 x + b^1 \right) + b^2
\end{align}
The output from each expert is weighted and combined based on the learned attention weight \( \operatorname{g}_{p,i} \) for each expert, yielding:
\begin{align}
    \mathbb{E}_{p}\left( x \right) = \sum_{i}{\operatorname{g}_{p,i}\left( x \right) \mathbb{E}_{p,i}\left( x \right)}
\end{align}
where \( \mathbb{E}_{p,i} \) denotes the output of the \( i^{th} \) expert in \( p^{th} \) pathway to input x. \( g_{p,i}\left( x \right) \) is a gating function that calculates the weight of each expert. The outputs from the motion, text, and cross-modal pathways are concatenated to produce the final representation:
\begin{align}
    \operatorname{Output} = \mathbf{W}_{proj}\left( [\mathbb{E}_{motion}; \mathbb{E}_{text}; \mathbb{E}_{cross-modal}] \right) + b_{proj}
\end{align}
Notably, all experts are activated simultaneously, with their contributions to each token dynamically adjusted by learned weights. This design benefits from building modality-sharing channels within each pathway, leveraging common features across datasets while supporting specialized learning.

\subsection{Training Strategy}
\paragraph{Stage 1: Train the MEVQ-VAE}
In the first stage, we train the MEVQ-VAE on raw motion data to learn a quantized representation of motion sequences. Following previous methods~\cite{zhang2023generating, pinyoanuntapong2024mmm}, the MEVQ-VAE model optimizes two losses, a reconstruction loss \( \mathcal{L}_{rec} \) and a commitment loss \( \mathcal{L}_{commit} \). The total loss can be expressed as:
\begin{align}
    \mathcal{L}_{q} = \mathcal{L}_{rec} + \beta \mathcal{L}_{commit}
\end{align}
where \( \beta \) is the hyper-parameter to balance the losses. Additionally, we utilize moving averages for codebook updates and resets for inactive codes~\cite{yu2022vectorquantized}.
\paragraph{Stage 2: Pretrain the GenM\(^3\)}
As shown in Fig. \ref{fig:pipeline} (b), in the pretraining phase, the GenM\(^3\) model is trained exclusively on motion data to establish a robust foundational representation of movement patterns. During this stage, text, motion, and cross-modal (motion-text) pathways all take motion tokens as inputs and utilize motion feature to form the prior knowledge that is useful in the next stage. We adopt a masked modeling approach on the motion token level to enhance GenM\(^3\)'s ability to reconstruct missing parts of motion data, following the method described in prior works~\cite{pinyoanuntapong2024mmm}. In each discrete motion token sequence, random indices are replaced by learnable special-purpose tokens \( [\operatorname{Mask}] \), encouraging the model to infer and regenerate these occluded motion parts. The objective is to minimize the negative log-likelihood of the predicted masked tokens conditioned on visible tokens:
\begin{align}
    \mathcal{L} = -\sum_{i \in \mathcal{M}}{\operatorname{log}P\left( x_i | x_{\setminus \mathcal{M}} \right)}
\end{align}
where \( \mathcal{M} \) denotes the set of masked indices, and \( x_{\setminus \mathcal{M}} \) represents the visible (non-masked) tokens in the sequence.

\paragraph{Stage 3: Text-conditional Training}
As shown in Fig. \ref{fig:pipeline} (c), in the text-conditional training phase, GenM\(^3\) is trained on both motion and textual data, enabling multimodal alignment between motion sequences and corresponding textual descriptions. During this stage, all pathways of the GenM\(^3\)--the motion, text, and cross-modal shared pathways--are active, allowing the model to fully leverage interactions within and across modalities. The motion pathway remains in a masked modeling framework, but depends on both textual and visible information.

\subsection{Inference Strategy}
During inference, all motion tokens are initialized as \( [\operatorname{Mask}] \) to form an empty sequence. Then, using parallel decoding, all tokens are generated simultaneously in each step. In subsequent iterations, tokens with lower confidence are re-masked and regenerated.

\begin{table*}[t]
\centering
\small
\caption{Comparison of Text-to-Motion on HumanML3D~\cite{guo2022humanml3d} (using our retrained evaluator on 30FPS motion data) and Zero-shot Text-to-Motion on IDEA400~\cite{lin2023motionx}. The highlight and underline represent the best and the second-best. Noting that GenM\(^3\) uses text pairs from HumanML3D for text-conditional training, while GenM\(^3\)\(^*\) denotes the use of all text-motion pairs from our integrated dataset.}
\begin{tabular*}{\textwidth}{@{\extracolsep{\fill}} ll ccc ccc @{}}
\toprule
\multirow{2}{*}{Dataset} & \multirow{2}{*}{Methods} & \multicolumn{3}{c}{R-Precision$\uparrow$} & \multirow{2}{*}{FID$\downarrow$} & \multirow{2}{*}{MMDist$\downarrow$} & \multirow{2}{*}{Diversity$\uparrow$} \\
\cmidrule(lr){3-5}
 &  & Top1 & Top2 & Top3 &  &  &  \\
\midrule
\multirow{5}{*}{HumanML3D} 
& Real & $0.492^{\pm.004}$ & $0.687^{\pm.003}$ & $0.785^{\pm.003}$ & $0.002^{\pm.000}$ & $2.982^{\pm.010}$ & $9.458^{\pm.090}$ \\
& T2M-GPT~\cite{zhang2023generating} & $0.487^{\pm.002}$ & $0.678^{\pm.003}$ & $0.770^{\pm.002}$ & $0.160^{\pm.006}$ & $3.083^{\pm.009}$ & $9.653^{\pm.065}$ \\
& MMM~\cite{pinyoanuntapong2024mmm} & $0.496^{\pm.003}$ & $0.686^{\pm.002}$ & $0.784^{\pm.002}$ & $0.110^{\pm.005}$ & $2.951^{\pm.009}$ & $9.484^{\pm.113}$ \\
& Ours GenM\(^3\)\(^*\) & $\underline{0.510}^{\pm.002}$ & $\underline{0.702}^{\pm.002}$ & $\underline{0.802}^{\pm.002}$ & $\underline{0.053}^{\pm.002}$ & $\underline{2.860}^{\pm.009}$ & $\underline{9.629}^{\pm.077}$ \\
& Ours GenM\(^3\) & $\boldsymbol{0.511}^{\pm.003}$ & $\boldsymbol{0.705}^{\pm.002}$ & $\boldsymbol{0.804}^{\pm.002}$ & $\boldsymbol{0.046}^{\pm.002}$ & $\boldsymbol{2.852}^{\pm.009}$ & $\boldsymbol{9.675}^{\pm.087}$ \\
\midrule
\multirow{5}{*}{IDEA400} 
& Real & $0.124^{\pm.003}$ & $0.221^{\pm.005}$ & $0.308^{\pm.006}$ & $0.001^{\pm.000}$ & $4.615^{\pm.004}$ & $6.001^{\pm.108}$ \\
& T2M-GPT~\cite{zhang2023generating} & $0.122^{\pm.003}$ & $0.219^{\pm.004}$ & $0.301^{\pm.004}$ & $7.947^{\pm.012}$ & $5.488^{\pm.034}$ & $7.636^{\pm.110}$ \\
& MMM~\cite{pinyoanuntapong2024mmm} & $0.126^{\pm.003}$ & $0.221^{\pm.005}$ & $0.307^{\pm.004}$ & $6.001^{\pm.010}$ & $4.980^{\pm.028}$ & $7.730^{\pm.103}$ \\
& Ours GenM\(^3\)\(^*\) & $\boldsymbol{0.136}^{\pm.004}$ & $\boldsymbol{0.244}^{\pm.002}$ & $\boldsymbol{0.338}^{\pm.003}$ & $\boldsymbol{4.232}^{\pm.008}$ & $\boldsymbol{4.520}^{\pm.012}$ & $\underline{7.750}^{\pm.101}$ \\
& Ours GenM\(^3\) & $\underline{0.134}^{\pm.003}$ & $\underline{0.241}^{\pm.005}$ & $\underline{0.335}^{\pm.003}$ & $\underline{4.430}^{\pm.060}$ & $\underline{4.732}^{\pm.007}$ & $\boldsymbol{7.851}^{\pm.131}$ \\
\bottomrule
\end{tabular*}
\label{tab:comparison_on_h3dnew_and_idea400_zero_shot}
\end{table*}

\begin{table}[t!]
\centering
\small
\renewcommand{\arraystretch}{1.25}
\setlength{\tabcolsep}{1pt}
\caption{Comparison of Text-to-Motion on HumanML3D~\cite{guo2022humanml3d} (using the evaluator trained on 20FPS motion data~\cite{guo2022generating}).}
\begin{tabular*}{\columnwidth}{@{\extracolsep{\fill}} c | c c c }
\toprule

Methods & FID\(\downarrow\) & R-Precision (Top3)\(\uparrow\) & Diversity\(\uparrow\) \\

\midrule

Real & 0.002 \(^{\pm.000}\) & 0.797\(^{\pm.002}\) & 9.503\(^{\pm.065}\) \\ 
MotionDiff~\cite{zhang2024motiondiffuse} & 0.630\(^{\pm.001}\) & 0.782\(^{\pm.001}\) & 9.410\(^{\pm.049}\) \\
MDM~\cite{tevet2023human} & 0.544\(^{\pm.044}\) & 0.611\(^{\pm.007}\) & 9.559\(^{\pm.086}\) \\
MLD~\cite{chen2023executing} & 0.473\(^{\pm.013}\) & 0.772\(^{\pm.002}\) & 9.724\(^{\pm.082}\) \\
T2M-GPT~\cite{zhang2023generating} & 0.141\(^{\pm.005}\) & 0.775\(^{\pm.002}\) & 9.722\(^{\pm.082}\) \\
EMDM~\cite{zhou2025emdm} & 0.112\(^{\pm.019}\) & 0.786\(^{\pm.006}\) & 9.551\(^{\pm.078}\)  \\
FineMoGen~\cite{zhang2023finemogen} & 0.151\(^{\pm.008}\) & 0.784\(^{\pm.002}\) & 9.263\(^{\pm.094}\) \\
MoMamba~\cite{zhang2025motion} & 0.281\(^{\pm.009}\) & 0.792\(^{\pm.002}\) & $\boldsymbol{9.871}^{\pm 0.084}$ \\
MoMask~\cite{guo2024momask} & \underline{0.045\(^{\pm.002}\)} & \textbf{0.807\(^{\pm.002}\)} & -  \\
MMM~\cite{pinyoanuntapong2024mmm} & 0.080\(^{\pm.003}\) & 0.794\(^{\pm.002}\) & 9.411\(^{\pm.058}\) \\
\midrule

MotionGPT~\cite{jiang2023motiongpt} & 0.232\(^{\pm.008}\) & 0.778\(^{\pm.002}\) & 9.528\(^{\pm.071}\) \\
MotionLLM~\cite{wu2024motionagent} & 0.491\(^{\pm.019}\) & 0.770\(^{\pm.002}\) &\underline{9.838\(^{\pm.244}\) }\\
MotionChain~\cite{Jiang2025motionchain} & 0.248\(^{\pm.009}\) & 0.790\(^{\pm.003}\) & 9.470\(^{\pm.075}\) \\
OMG~\cite{liang2024omg} & 0.381\(^{\pm.008}\) & 0.784\(^{\pm.002}\) & 9.657\(^{\pm.085}\) \\

\midrule

Our GenM\(^3\)\(^*\) & 0.046\(^{\pm.002}\) & 0.790\(^{\pm.003}\) & 9.346\(^{\pm.076}\) \\ 
Our GenM\(^3\) & \textbf{0.035\(^{\pm.002}\)} & \underline{0.795\(^{\pm.002}\)} & 9.341\(^{\pm.080}\) \\ 
\bottomrule
\end{tabular*}
\label{tab:comparisons_on_humanml3d}
\vspace{-0.5cm}
\end{table}

\section{Integration of Motion Data}
\paragraph{Data Standardization}
To train a motion learning backbone capable of comprehensive representation, we collected a diverse set of highly accurate motion capture datasets. Given the disparities across datasets in terms of format (such as spatial coordinates versus SMPL parameters), body shape of performers, joint count, and frame rate, we standardized all datasets accordingly. Specifically, we normalized the joint count to 22, aligning with the HumanML3D~\cite{guo2022humanml3d} standard, and retargeted each dataset to a default human skeletal template. To prevent issues with undersampling, we set the frame rate to 30 fps. 

\paragraph{Text Modality}
Our dataset includes HumanML3D~\cite{guo2022humanml3d}, HuMMan~\cite{cai2022humman}, and HIMO~\cite{lv2025himo}, each with textual descriptions for every motion sequence. However, motion data inherently differs from text in structure and contextual relationships, and there are distribution differences in motion/text data across different datasets. HumanML3D emphasizes holistic human movement. HuMMan provides detailed motion patterns at the joint level. While HIMO focuses on interactions between the human upper body and objects. Additionally, the BABEL dataset provides action categories for each sub-segment. We divide the sequences based on sub-segment duration and generated sequence descriptions for each complete segment using ChatGLM~\cite{glm2024chatglm}. In order to further increase the diversity of text data, we manually divide sequences and label motion descriptions on the human-body-object interaction dataset HDMI.

\paragraph{Sequence Segmentation} During pretraining, for datasets without predefined or suitable sequence segmentation (e.g., sequences over 10 seconds), we sample training sequences with overlapping windows, ensuring overlap rates were kept below 25\% to maintain diversity. For text-conditional generative training, we follow the original segmentation provided by each dataset or our manual partitioning.
\begin{figure*}[t!]
    \centering
    \includegraphics[width=0.99\textwidth]{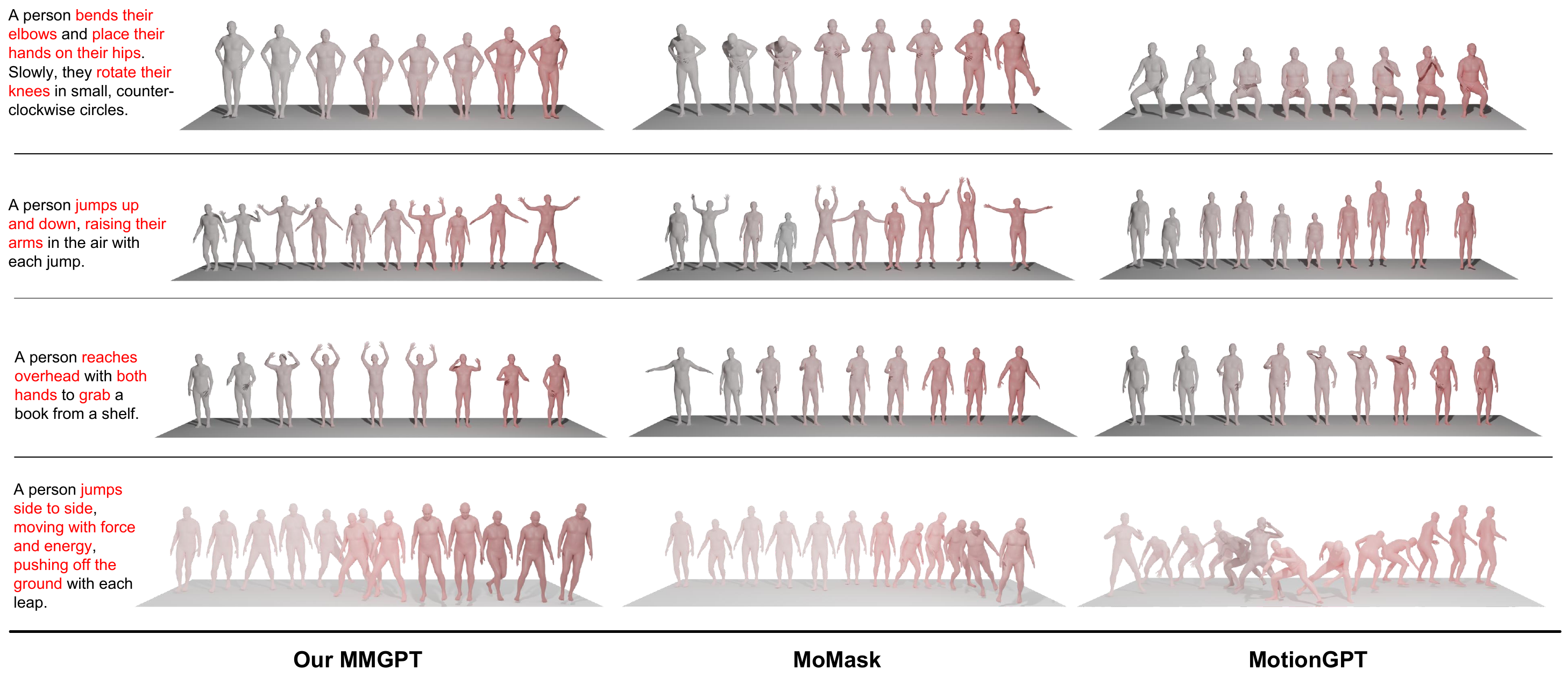}
    \vspace{-0.25cm}
    \caption{Motion generation results based on text inputs.} 
    \label{fig:generation}
    \vspace{-0.25cm}
\end{figure*}

\section{Experiments}
This section provides a comprehensive evaluation of our proposed GenM\(^3\) framework. In Sec. \ref{5.1}, we introduce the datasets and experimental setup. Sec. \ref{5.2} presents a comparative analysis of our method against \textit{state-of-the-art} methods. In Sec. \ref{5.3}, we conduct ablation studies to investigate the contributions of individual components. We provide more information in the supplemental.

\begin{figure}[t!]
    \centering
    \includegraphics[width=0.45\textwidth]{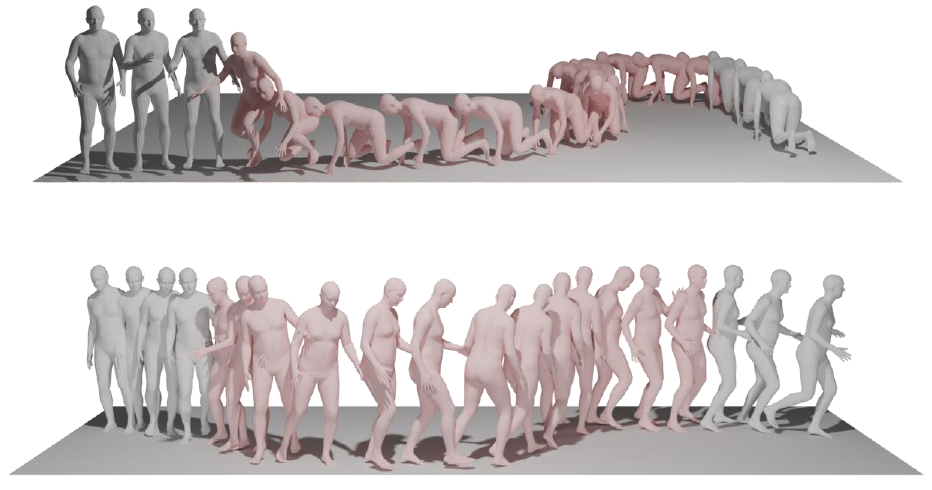}
    \vspace{-0.25cm}
    \caption{Visualizations of motion in-between.} 
    \label{fig:in-between}
    \vspace{-0.5cm}
\end{figure}

\subsection{Datasets and Implementation}
\label{5.1}
\paragraph{Datasets for Pretraining and Text-conditioned Training}
We collected and processed a total of 11 motion capture datasets~\cite{araujo2023circle, zhang2022egobody, taheri2020grab, lv2025himo, zhao2024imhoi, punnakkal2021babel, guo2022humanml3d, cai2022humman, mehta2017monocular, liu2024emage, liang2024intergen}, aligning each to the HumanML3D dataset's format for consistency in training and evaluation. We use all the motion data for pretraining. For text-conditional training, we used datasets that include text modality: HumanML3D~\cite{guo2022humanml3d}, HuMMan~\cite{cai2022humman}, HIMO~\cite{lv2025himo}, BABEL~\cite{punnakkal2021babel}, and IMHD~\cite{zhao2024imhoi}.

\paragraph{Implementation Details}
Our model is trained on Nvidia RTX-4090 GPUs. In MEVQ-VAE, motion sequences are downsampled by a factor of 4, and both encoder and decoder use a four-expert multi-expert architecture. The codebook has 8192 entries of 32 dimensions, with the loss weight \( \beta \) set to 1. The MMT contains 18 layers: the first nine layers follow standard attention, while the latter nine layers implement multi-path transformer with multiple experts. For pretraining and text-conditional training, we set the batch size to 160 and use AdamW optimizer to train the model for 120,000 iterations with a learning rate of \( 2\times10^{-4} \) following a warm-up and cosine annealing decay strategy.

\subsection{Comparison}
\label{5.2}
\paragraph{Quantitative Comparison}
We first perform a quantitative evaluation on the HumanML3D~\cite{guo2022humanml3d}. We report the results evaluated by the evaluator from prior works~\cite{guo2022humanml3d, pinyoanuntapong2024mmm} in Tab.~\ref{tab:comparisons_on_humanml3d}. However, this evaluator was trained on 20 FPS motion samples, which may not be suitable to evaluate on our constructed dataset. Thus, we re-trained the evaluator following the prior settings~\cite{guo2022humanml3d} and trained MMM and T2M-GPT on 30FPS HumanML3D for fair comparisons. The results are shown in Tab.~\ref{tab:comparison_on_h3dnew_and_idea400_zero_shot}. Following the standard evaluation protocols, we report the results from 20 repeated trials, with each result shown as the mean with a 95\% confidence interval. We trained two versions of GenM\(^3\): GenM\(^3\), using only HumanML3D text pairs for text-conditional training, and GenM\(^3\)\(^*\), using all motion-text data. GenM\(^3\) performs better on HumanML3D, but GenM\(^3\)* exhibits stronger generalization ability (will be discussed later). It can be seen that Our method outperforms other approaches in FID and achieves comparable performances in other metrics.

\paragraph{Qualitative Comparison}
Fig.~\ref{fig:generation} presents qualitative results showcasing the generated motions. We used different types of text as conditions, including: 1) descriptions based on body joint movements (e.g., "place their hands on their hips"); and 2) descriptions based on motion types (e.g., "jump"). Due to the limited diversity and quantity of text in the HumanML3D dataset, methods trained on this dataset often fail to generate effective motion sequences for some simple and common action descriptions. Our approach, however, demonstrates a clear advantage to a wide range of descriptions and high-quality generations.

\begin{table}[t]
    \centering
    \renewcommand{\arraystretch}{0.9} 
    \setlength{\tabcolsep}{4pt}      
    \scriptsize
    \caption{Comparison of different VQ methods.}
    \begin{tabular}{@{}c|cccccc@{}}
        \toprule
        Method & VQ & MEVQVAE & RVQ~\cite{guo2024momask} & MERVQ & G-RVQ~\cite{yang2023hifi} & FSQ~\cite{mentzer2024finite}  \\
        \midrule
        FID    & 0.098 & 0.048   & 0.043 & 0.032  & 0.045        & 0.057 \\
        \bottomrule
    \end{tabular}
    \label{tab:quantizer}
\end{table}

\paragraph{Comparison of VQ Method} 
Tab.~\ref{tab:quantizer} compares different quantization methods, including RVQ~\cite{guo2024momask}, G-RVQ~\cite{yang2023hifi} and FSQ~\cite{mentzer2024finite}. Our multi-expert design adapts to diverse dataset distributions, substantially outperforming standard VQ. RVQ achieves the best scores but requires an extra Residual Transformer to model residuals~\cite{guo2024momask}. Integrating our Multi-Expert into RVQ (MERVQ) further boosts performance, demonstrating its generalizability.

\begin{table}[t!]
\small
\centering
\renewcommand{\arraystretch}{1.25}
\setlength{\tabcolsep}{8.5pt}
\caption{Results of motion in-betweening. R. means R-precision. Div. means Diversity.}
\begin{tabular}{lc|ccc}
\toprule
Type & Method & R.(Top 3)$\uparrow$ & FID$\downarrow$ & Div.$\uparrow$ \\ 
\midrule
\multirow{2}{*}{Prefix} 
& MMM~\cite{pinyoanuntapong2024mmm} & 0.780 & 0.098 & 9.432 \\
& Ours &  $ \boldsymbol{0.797}$ &$\boldsymbol{0.040}$ &$\boldsymbol{ 9.583}$\\

\midrule
\multirow{2}{*}{Suffix} 
& MMM~\cite{pinyoanuntapong2024mmm} & 0.788 & 0.142 & 9.455 \\
& Ours &  $ \boldsymbol{0.800}$  & $\boldsymbol{0.065}$  & $\boldsymbol{ 9.537}$  \\
\midrule
\multirow{2}{*}{Infix} 
& MMM~\cite{pinyoanuntapong2024mmm} & 0.784 & 0.133 & 9.398 \\
& Ours &  $ \boldsymbol{0.803}$  & $\boldsymbol{0.056}$  & $\boldsymbol{ 9.506}$  \\
\bottomrule
\end{tabular}
\label{tab:motion-in-between}
\vspace{-0.25cm}
\end{table}

\paragraph{Motion Completion}
We conducted comparative experiments with MMM~\cite{pinyoanuntapong2024mmm} on the motion completion task on the HumanML3D test set, evaluating three distinct modes: prefix-based (generating the last 50\% based on the first 50\% of observed motion), suffix-based (generating the first 50\% based on the last 50\% of observed motion), infix-based (generating the middle 50\% based on the initial and final 25\% of observed motion) completion. As shown in the Tab.~\ref{tab:motion-in-between}, our method achieves better metrics especially the FID. Fig.~\ref{fig:in-between} visualizes some of the motion completion results, where it can be seen that our method excels in both overall semantic coherence and motion transition.

\paragraph{Zero-shot Inference}
We tested the zero-shot capability of GenM\(^3\) on IDEA400 dataset. GenM\(^3\)* refers to the model trained on all motion-text pairs in our integrated dataset. As can be seen in Tab.~\ref{tab:comparison_on_h3dnew_and_idea400_zero_shot}, thanks to the strong representational power learned during pre-training, the GenM\(^3\) model trained only on HumanML3D still exhibits robust zero-shot reasoning ability. GenM\(^3\)*, trained on more motion-text pairs, shows even stronger generalization ability.

\subsection{Ablation Study}
\label{5.3}
\begin{figure}[t!]
    \centering
    \includegraphics[width=0.46\textwidth]{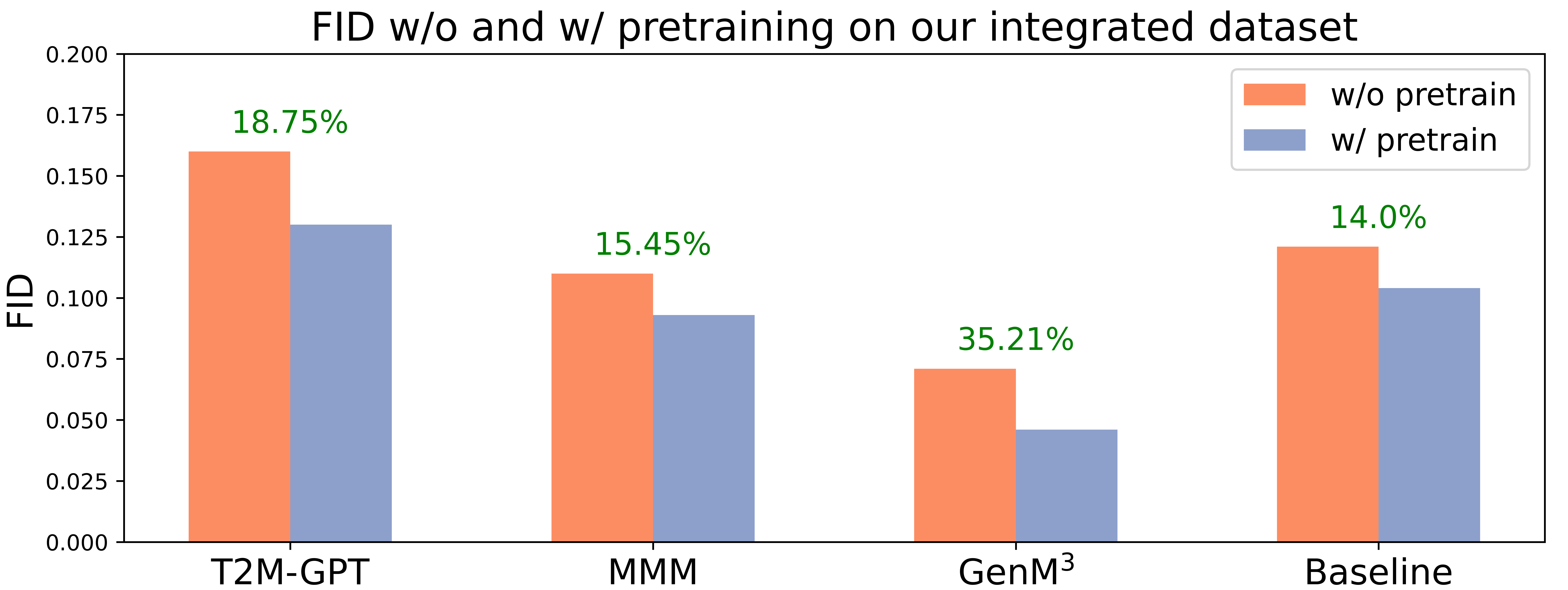}
    \vspace{-0.25cm}
    \caption{The impact of pretraining on different methods on our integrated dataset.} 
    \label{fig:pretrain}
    \vspace{-0.35cm}
\end{figure}

\paragraph{Pretraining Improvement}
We conducted experiments by pretraining MMM~\cite{pinyoanuntapong2024mmm}, T2M-GPT~\cite{zhang2023generating} and our baseline (refers to our method without the Multi-path, Multi Expert, and Motion Descriptor components) on our integrated dataset in Fig.~\ref{fig:pretrain}. It can be seen that pretraining on our integrated dataset yields the greatest improvement (35.21\%) for our method. This can be attributed to the multi-expert design in our MEVQ-VAE, which enables the model to adapt to the heterogeneity of various datasets. Moreover, the Motion Pathway in MMT leverages multiple experts to simultaneously learn motion patterns from different datasets.

\paragraph{Effective of Multi-path Transformer in GenM\(^3\)}
We conducted ablation experiments on the text and cross-modal branches in GenM\(^3\), with the results shown in Tab.~\ref{tab:mmtrans}. As can be seen, when both branches are used simultaneously, the model is able to more effectively learn the features both within and across modalities.

\begin{figure}[t]
    \centering
    \includegraphics[width=0.465\textwidth]{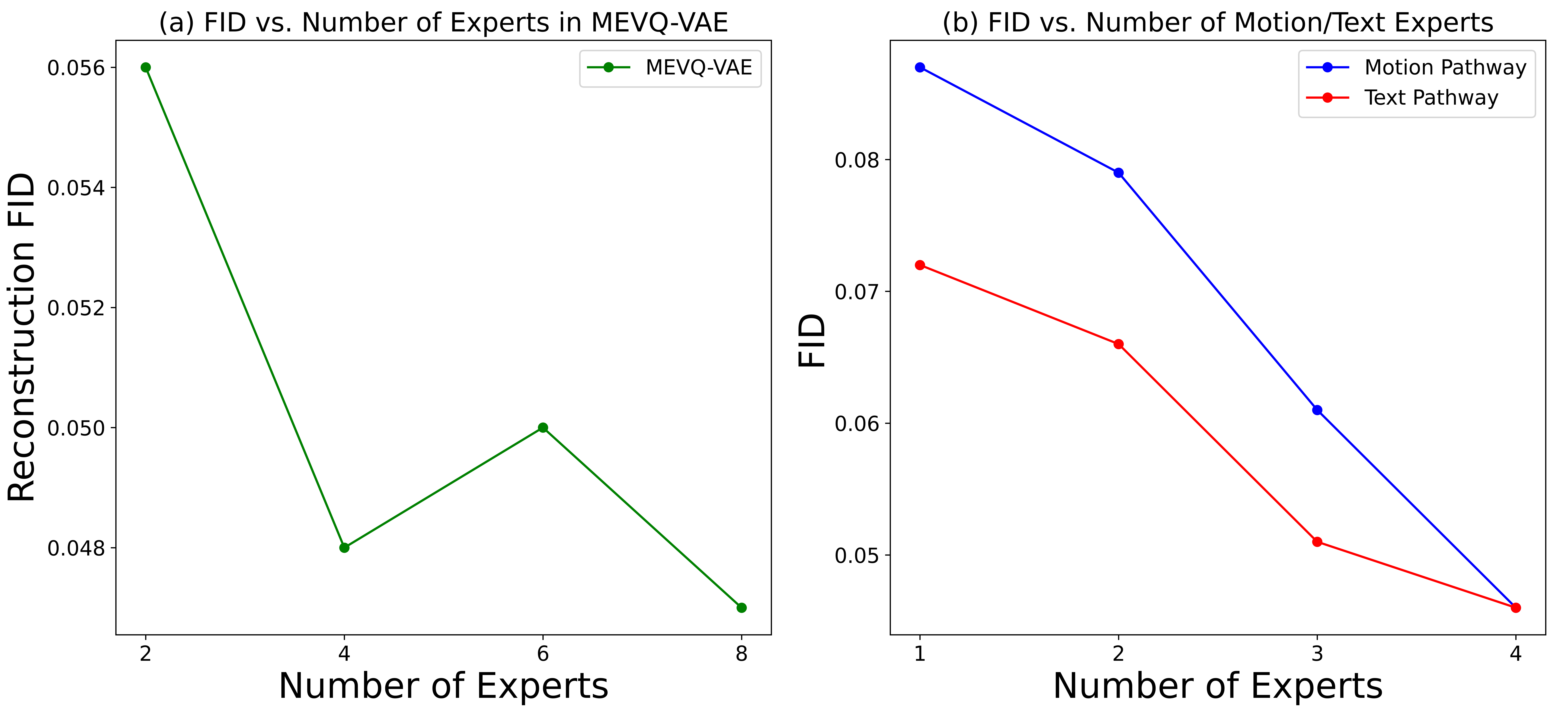}
    \vspace{-0.25cm}
    \caption{The impact of the number of experts in MEVQ-VAE on reconstruction performance.} 
    \label{fig:experts}
\end{figure}

\paragraph{Impact of Expert Count}
We analyzed the effect of expert numbers on three components. In MEVQ-VAE (Fig.~\ref{fig:experts} (a)), using eight experts produced the best reconstruction performance. However, due to computational costs, we set up four experts to get nearly the same result. In the motion and text pathways (Fig.~\ref{fig:experts} (b)), increasing the number of experts enhances the model's ability to handle a broader range of data distributions (e.g. diverse textual descriptions and diverse motion patterns present across datasets). 

\begin{table}[t]
\small
\centering
\caption{Comparisons of dense MoE and sparse MoE.}
\vspace{-0.15cm}
\setlength{\tabcolsep}{6.5pt}
\begin{tabular}{c|ccc}
\toprule
 & FID$\downarrow$ & R-Precision (Top3)$\uparrow$ &Diversity$\uparrow$\ \\ 
 \midrule
Sparse MoE & 0.058 & 0.799 &  $ \boldsymbol{ 9.676}$   \\
Dense MoE & $ \boldsymbol{0.046}$ & $ \boldsymbol{ 0.804}$  & 9,675 \\
\bottomrule
\end{tabular}
\label{tab:moe}
\end{table}

\paragraph{Dense MoE \textit{v.s.} Sparse MoE}
We conducted comparisons of dense MoE (fully activated) and sparse MoE (partially activated) in the motion pathway. As shown in Tab.~\ref{tab:moe}, fully activated experts better capture commonalities across datasets and adapt different dataset from a higher dimension. While Sparse MoE may overlook some shared patterns and failed adapt the data in a narrower action space.

\begin{table}[t]
\small
\centering
\caption{The impact of the multiway transformer in GenM\(^3\) on generation performance.}
\label{tab:mmtrans}
\vspace{-0.15cm}
\setlength{\tabcolsep}{9pt}
\begin{tabular}{ccc|cc}
\toprule
motion & text & cross-modal & FID$\downarrow$ & Diversity$\uparrow$\\ 
\midrule
\checkmark & - & - & 0.058 &  $\boldsymbol{ 9.400} $  \\
\checkmark & \checkmark & - & 0.045 & 9.282 \\
\checkmark & - & \checkmark &$\underline{0.044} $  & $\underline{9.344}$ \\
\checkmark & \checkmark & \checkmark & $\boldsymbol{0.035} $  & 9.341 \\
\bottomrule
\end{tabular}
\vspace{-0.15cm}
\end{table}

\section{Conclusion}
In this paper, we propose GenM\(^3\), which consists of a multi-expert VQ-VAE to learn the unified discrete motion representation, and a Multi-path Motion Transformer that enhances intra-modal representations by employing dedicated pathways for each modality, with densely activated experts that capture the inherent variations, and bolsters inter-modal alignment through a shared text-motion pathway. We integrated and unified 11 high-quality motion datasets to support the training of GenM\(^3\). Experimental results show that our method achieves state-of-the-art performance on the Text-to-Motion task while demonstrating remarkable generalization capabilities. In the future, we plan to further expand this dataset to train a more generalizable motion backbone for the research community.
\newpage
\section{Acknowledgment}
This work is partially supported by the Guangzhou-HKUST(GZ) Joint Funding Program (No. 2025A03J3656). 
{
    \small
    \bibliographystyle{ieeenat_fullname}
    \bibliography{main}
}
\clearpage
\setcounter{page}{1}
\maketitlesupplementary

In this supplementary material, we provide more information that could not be included in the main manuscript because of space limit. We first provide more details about our constructed dataset and implementations in Sec.~\ref{sec.7} and~\ref{sec.8}. More results and ablation studies are conducted in Sec.~\ref{sec.9} Sec.~\ref{sec.10} shows the visualization results of generated motion corresponding to different text inputs. Finally, we discuss limitations of our proposed GenM\(^3\) in Sec.~\ref{sec.11}.

\begin{table}[t]
  \centering
  \small
  \renewcommand{\arraystretch}{1.25}
  \setlength{\tabcolsep}{6.5pt}
  \caption{Dataset statistics. We calculate the total number of frames and duration and the number of textual descriptions for each dataset.}
  \label{tab:datasets}
    \begin{tabular}{c | ccc}
      \toprule
      Dataset & Frame(M) & Duration(h) & Text \\
      \midrule
      BABEL~\cite{punnakkal2021babel}       & 2.25  & 20.84  & 9,742  \\
      BEATv2~\cite{liu2024emage}           & 9.43  & 87.29  & –      \\
      CIRCLE~\cite{araujo2023circle}       & 1.07  & 9.91   & –      \\
      EgoBody~\cite{zhang2022egobody}      & 0.40  & 3.70   & –      \\
      GRAB~\cite{taheri2020grab}           & 0.40  & 3.74   & –      \\
      HIMO~\cite{lv2025himo}               & 1.14  & 10.59  & 2,711  \\
      HumanML3D~\cite{guo2022humanml3d}    & 6.09  & 56.42  & 29,226 \\
      HuMMan~\cite{cai2022humman}          & 0.69  & 6.35   & 6,264  \\
      IMHD~\cite{zhao2024imhoi}            & 0.13  & 1.21   & 308    \\
      MPI-INF-3DHP~\cite{mehta2017monocular}& 0.13  & 1.22   & –      \\
      InterHuman~\cite{liang2024intergen}  & 2.02  & 18.69  & –      \\
      \midrule
      Total                                &23.75  &219.96  &48,251  \\
      \bottomrule
    \end{tabular}%
\end{table}

\section{Details of the Dataset}
\label{sec.7}
We collected and processed 11 high-quality motion capture datasets, including CIRCLE~\cite{araujo2023circle}, Egobody~\cite{zhang2022egobody}, GRAB~\cite{taheri2020grab}, HIMO~\cite{lv2025himo}, IMHD~\cite{zhao2024imhoi}, BABEL~\cite{punnakkal2021babel}, HumanML3D~\cite{guo2022humanml3d}, HuMMan~\cite{cai2022humman}, MPI-INF-3DHP~\cite{mehta2017monocular}, InterHuman~\cite{liang2024intergen} and BEATv2~\cite{liu2024emage}. All datasets were processed at 30 fps, with the duration of each sample ranging from 2 to 10 seconds. Tab.~\ref{tab:datasets} illustrates the number of frames included in each dataset after processing.

\paragraph{Pose Representation}
We standardized all datasets to align with the format of the HumanML3D dataset. A pose \( p \) is represented  as a tuple consisting of multiple components: \( \left( r^{a}, r^{x}, r^{z}, r^{y}, \mathbf{j}^{p}, \mathbf{j}^{v}, \mathbf{j}^{r}, \mathbf{c}^{f} \right) \). Here, \( r^{a} \in \mathbb{R} \) represents the root angular velocity along the Y-axis, while \( \left( r^{x}, r^{z} \in \mathbb{R} \right) \) describe the root's linear velocities on the XZ-plane. The root height is denoted as \( r^{y} \in \mathbb{R} \). The joint-related attributes include \( \mathbf{j}^{p} \in \mathbb{R}^{3j}, \mathbf{j}^{v} \in \mathbb{R}^{3j}, \mathbf{j}^{r} \in \mathbb{R}^{6j} \), which respectively correspond to the positions, velocities, and rotations of joints in the root space, where \( j \) indicates the total number of joints. Additionally, \( \mathbf{c}^{f} \in \mathbb{R}^4 \) is a set of binary features, derived by thresholding the velocities of the heel and toe joints to highlight ground contact points.

\begin{figure}[t]
    \centering
    \includegraphics[width=0.43\textwidth]{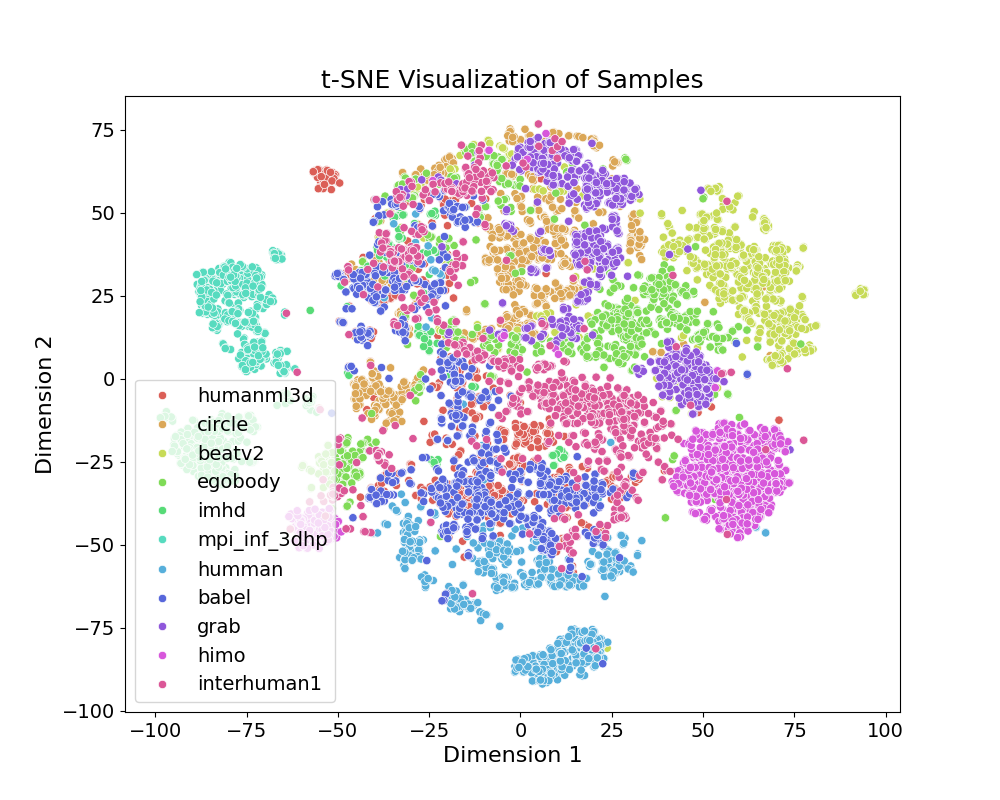}
    \caption{Visualization of data distribution after dimensionality reduction using T-SNE algorithm.} 
    \label{fig:datasets}
\end{figure}

\paragraph{Dataset Distribution}
We randomly sampled 1,000 motion sequences from each dataset and reduced their dimensionality to two using the T-SNE algorithm. Fig. ~~\ref{fig:datasets} presents a visualization of the data distribution after dimensionality reduction, which clearly demonstrates significant differences in data distribution across the datasets.

\paragraph{Text Label}

HumanML3D, HuMMan, and HIMO datasets provide motion sequences paired with corresponding text labels. For these datasets, we adhered to their original splits and utilized the provided textual annotations. HDMI is a dataset focused on human-object interaction. For this dataset, we manually segmented the motion sequences based on video content. During the text annotation process, we described the human motions and the specific interactions with objects in detail. For example: \textit{"The person is holding a pan with their left hand at chest level and a spatula with their right hand, repeatedly stirring. Simultaneously, they sway their body from side to side, with the left foot half a step ahead of the right, and their head looking at the pan."}

The BABEL dataset provides action category information for each subsequence. To process this dataset, we first divided the overlapping subsequences into motion segments ranging from 2 to 10 seconds in duration based on their lengths. Subsequently, we used ChatGLM~\cite{glm2024chatglm} to automatically generate a concise textual description of the actions in each sequence, arranging the action categories in temporal order. The prompt for ChatGLM was as follows:

\textcolor[rgb]{0.5,0.7,0.8}{\textit{The user will give you a sequence of words. Please use these words to form one or a few concise sentences in English to describe the person's actions. Do not add any unrelated descriptions or overly elaborate embellishments. For example, if given the words walk, scoop, place, turn, walk, output: The person is walking over, scooping up the item, placing it down, turning, and walking away. If the direction is not clear, do not add directional modifiers, but if direction words like "turn" are present, you can add directional descriptions, such as "walk over" or "walk away."}}

This process ensures accurate and consistent textual descriptions across datasets.

\section{Implementation Details}
\label{sec.8}
Our model is implemented using PyTorch 2.0 and trained on Nvidia RTX-4090 GPUs. The architecture consists of two main components: MEVQ-VAE and Multi-path Motion Transformer. In MEVQ-VAE, motion sequences are downsampled by a factor of four during discretization, reducing temporal resolution. Both the encoder and decoder employ a multi-expert architecture with a default of four experts. The codebook is designed with 8,192 entries, each of 32 dimensions, providing a rich discrete representation space for motion features. To address codebook collapse, where large codebooks risk underutilization, we adopt a factorized code approach~\cite{yu2022vectorquantized}, decoupling code lookup and embedding, and use moving averages for updates while resetting inactive codes to enhance utilization. During training, the loss weight parameter \( \beta \) is set to 1, balancing reconstruction and commitment losses to ensure effective encoding.

The Multi-path Motion Transformer comprises of 18 layers: the first nine layers utilize standard attention mechanisms for initial feature extraction, while the latter nine layers implement a multiway transformer with multiple experts. All attention computations use 16 heads, each with a dimension of 64. For both pretraining and text-conditional training, we use a batch size of 160 and the AdamW optimizer, training the model for 120,000 iterations. The learning rate is set to 0.0002, following a warm-up and cosine annealing decay strategy to facilitate rapid adaptation in early training.

\section{More Results}
\label{sec.9}

\begin{table}[t]
    \centering
    \caption{Inference speed per sample (on a RTX4090 GPU) with different iteration settings during masked decoding.}
    \label{tab:efficiency}
    \scriptsize
    \setlength{\tabcolsep}{6.5pt}
    \begin{tabular}{l|cccc}
        \toprule
        Method & FID & Infer. time (ms) & R-Precision & Diversity \\
        \midrule
        GenM\(^3\) (iter.=1) & 2.242 & \textbf{23.73} & 0.679 & 8.477 \\
        GenM\(^3\) (iter.=5) & 0.054 & 67.49 & 0.785 & 9.375 \\
        GenM\(^3\) (iter.=10) & \textbf{0.046} & 113.22 & 0.804 & 9.675 \\
        GenM\(^3\) (iter.=15) & 0.060 & 188.12 & \textbf{0.805} & \textbf{9.719} \\
        \midrule
        T2M-GPT & 0.160 & 108.03 & 0.770 & 9.653 \\
        \bottomrule
    \end{tabular}
\end{table}

\paragraph{Inference Speed}
During decoding, all tokens are decoded in parallel at each iteration. Tab.~\ref{tab:efficiency} shows how iteration count (10 for default settings) relates to inference time, and compares this with the AR model T2M-GPT~\cite{zhang2023generating}. 

\begin{table}[t]
    \centering
    \scriptsize
    \caption{Results on HumanML3D under varying data proportions.}
    \label{tab:results}
    \setlength{\tabcolsep}{9.5pt}
    \begin{tabular}{c|cccc}
        \toprule
        Data Proportion & 0\% & 35\% & 70\% & 100\% \\
        \midrule
        FID             & 0.071 & 0.060 & 0.050 & 0.046 \\
        \bottomrule
    \end{tabular}
    \label{tab:data_proportion}
\end{table}

\paragraph{Influence of Dataset Size}
We have evaluated our backbone pre-training on varying proportions of the dataset (HumanML3D dataset is fully used). We can conclude that the quality of the generation scales with the size of dataset (see Tab.~\ref{tab:data_proportion}). We believe that integrating additional motion data can unlock further model potential, which will be the focus of our future work.

\begin{table}[t]
    \centering
    \scriptsize
    \caption{Results on HumanML3D (annotated by FineMotion).}
    \label{tab:finemotion}
    \begin{tabular}{lcccccc}
        \toprule
        Method & FID & R.Top1 & R.Top2 & R.Top3 & MMDist. & Diversity \\
        \midrule
        MMM & 24.7 & 0.060 & 0.113 & 0.159 & 5.85 & 6.01 \\
        GenM\(^3\) & 17.2 & 0.072 & 0.135 & 0.191 & 5.30 & 6.00 \\
        GenM\(^3\)\(^*\) & \textbf{12.9} & \textbf{0.087} & \textbf{0.154} & \textbf{0.218} & \textbf{5.01} & \textbf{6.25} \\
        \bottomrule
    \end{tabular}
\end{table}

\paragraph{Zero-shot Generation}
We added comparisons on unseen HumanML3D annotations from FineMotion\footnote{\url{https://github.com/BizhuWu/FineMotion }} (see Tab.~\ref{tab:finemotion}) to further prove GenM\(^3\)\(^*\)'s generalization ability. Since the re-annotated descriptions differ significantly from those in the original HumanML3D, the zero-shot generation performance is generally not ideal. However, our method achieves better results compared to MMM~\cite{pinyoanuntapong2024mmm}. By comparing the results of GenM\(^3\)\(^*\) and GenM\(^3\), it can be concluded that training on a larger dataset (with more text annotations and more motion data) improves the model's generalization ability.

\begin{figure*}[t]
    \centering
    \includegraphics[width=1\textwidth]{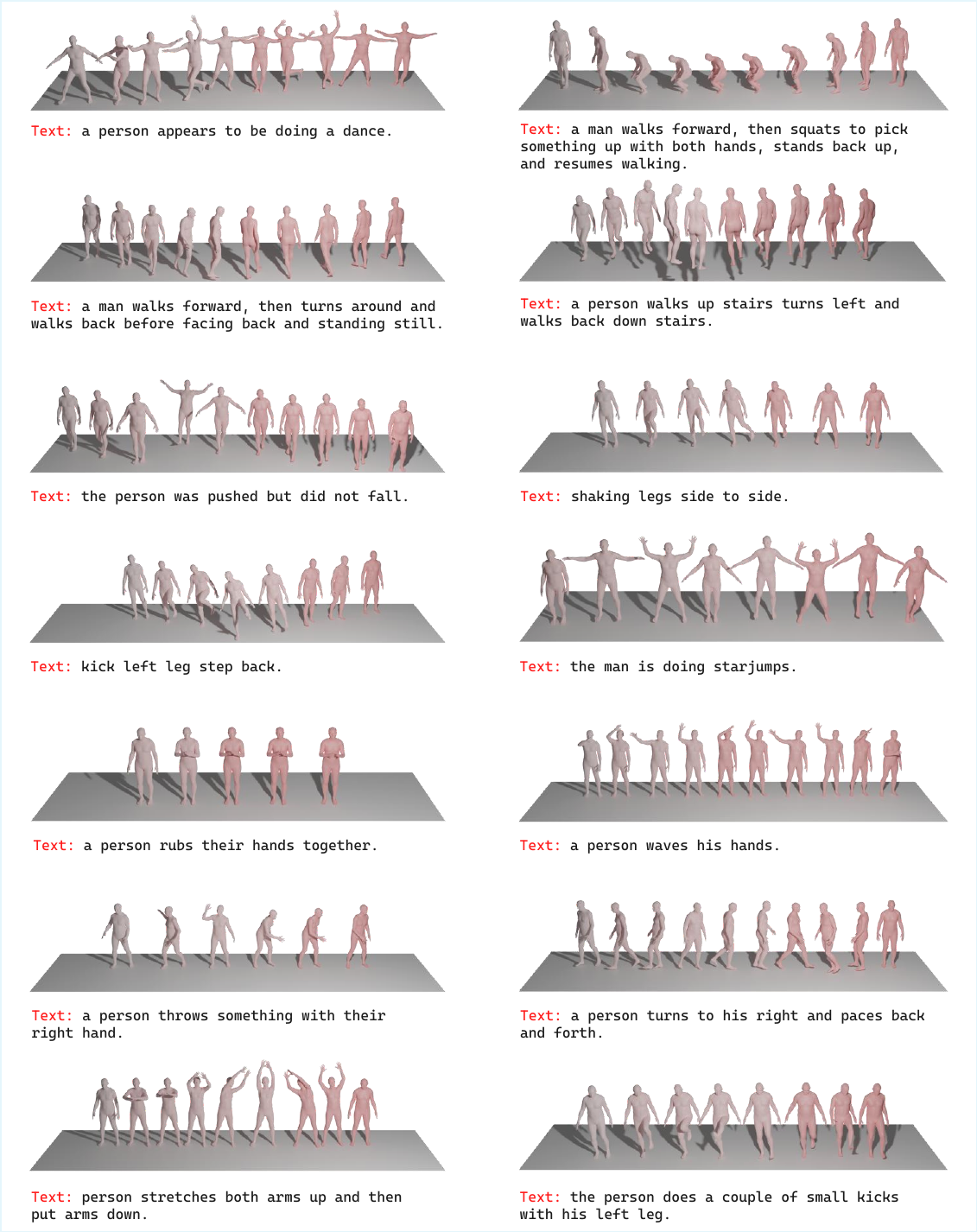}
    \caption{Visualizations of generated motion samples on HumanML3D~\cite{guo2022humanml3d} test set.} 
    \label{fig:vis_h3d}
\end{figure*}

\begin{figure*}[t]
    \centering
    \includegraphics[width=1\textwidth]{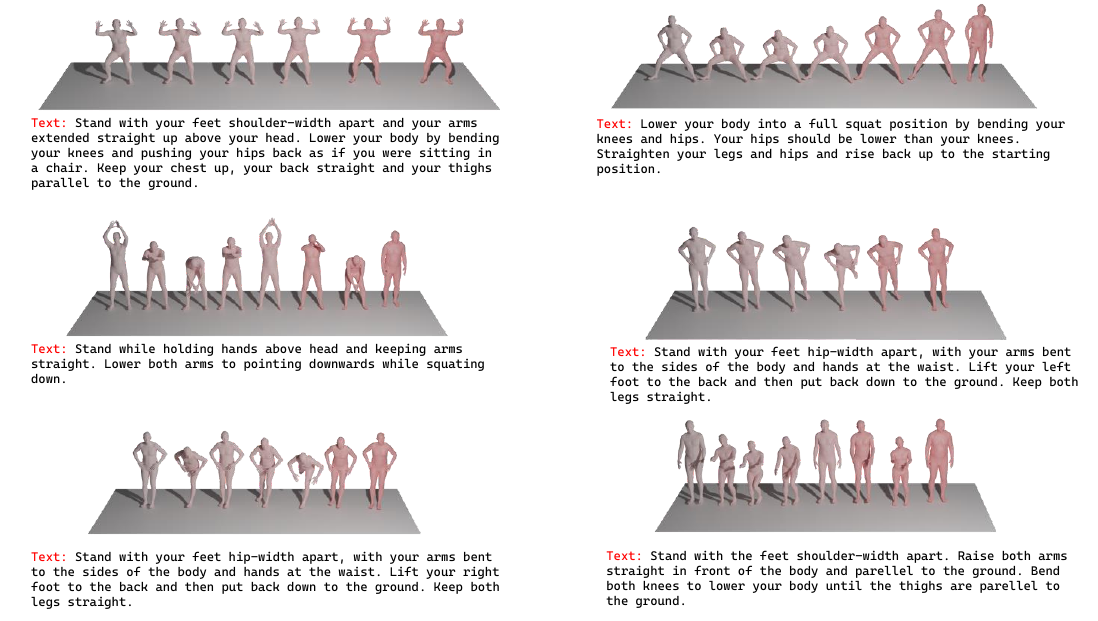}
    \caption{Visualizations of generated motion samples on HuMMan~\cite{cai2022humman} test set.} 
    \label{fig:vis_humman}
\end{figure*}

\begin{figure*}[t]
    \centering
    \includegraphics[width=1\textwidth]{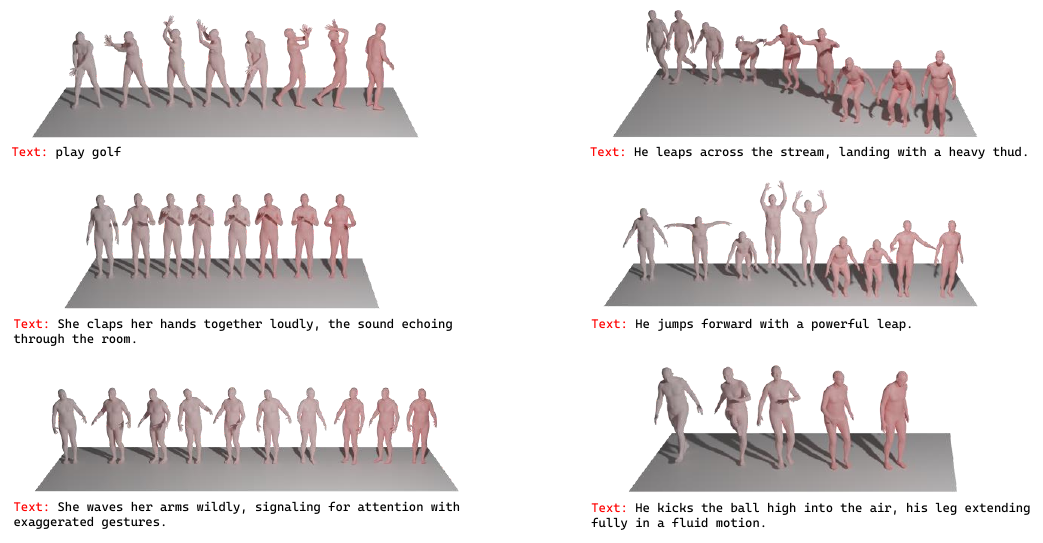}
    \caption{Visualizations of generated motion samples on GPT-generated textual descriptions.} 
    \label{fig:vis_others}
\end{figure*}

\section{Visualization}
\label{sec.10}
We provide more qualitative results from motion generation experiments using text inputs from the HumanML3D test set (Fig.~\ref{fig:vis_h3d}), the HuMMan test set (Fig.~\ref{fig:vis_humman}), and GPT-generated textual descriptions (Fig.~\ref{fig:vis_others}).

\section{Limitation}
\label{sec.11}
Our approach has two main limitations. First, although we enhanced the dataset by adding text labels for various motion types, the number of these labels remains limited. As a result, our method struggles to generate accurate motions for descriptions that fall outside the dataset's text distribution. In the future, we plan to explore ways to leverage additional text-motion pair data or integrate video-text pairs to enable the model to better comprehend diverse textual descriptions. Second, our current approach focuses primarily on body motion generation, overlooking the generation of full-body motions, such as fine-grained movements of fingers and facial expressions. In future work, we plan to collect and process more comprehensive datasets to enable MMGPT to generate more detailed and precise motions for all human joints.

\end{document}